# A Review of Image Mosaicing Techniques

**Dushyant Vaghela[1]**
Researcher at M.Tech Computer Engineering Dept.
R.K. University
Rajkot, India
Vaghela.dushyant@gmail.com

**Prof. Kapildev Naina[2]**
M.Tech Computer Engineering Dept.
R.K. University
Rajkot, India
Kapildev.naina@rku.ac.in

*Abstract— Image Mosaicing is a method of constructing multiple images of the same scene into a larger image. The output of the image mosaic will be the union of two input images. Image-mosaicing algorithms are used to get mosaiced image. Image Mosaicing processed is basically divided in to 5 phases. Which includes; Feature point extraction, Image registration, Homography computation, Warping and Blending if Image. Various corner detection algorithm is being used for Feature extraction. This corner produces an efficient and informative output mosaiced image. Image mosaicing is widely used in creating 3D images, medical imaging, computer vision, data from satellites, and military automatic target recognition.*

## I. INTRODUCTION

Image Mosaicing technology is becoming more and more popular in the fields of image processing, computer graphics, computer vision and multimedia. It is widely used in daily life by stitching pictures into panoramas or a large picture which can display the whole scenes vividly. For example, it can be used in virtual travel on the internet, building virtual environments in games and processing personal pictures. In Image Mosaicing is firstly divided into (usually equal sized) rectangular sections, each of which is replaced with another photograph that matches the target photo. When viewed at low magnifications, the individual pixels appear as the primary image, while close examination reveals that the image is in fact made up of many hundreds or thousands of smaller images.

In image mosaicing two input images are taken and this images are fused to form a single large image. This merged single image is the output mosaiced image. The first step in Image Mosaicing is feature extraction. In feature extraction, features are detected in both input images. Image registration refers to the geometric alignment of a set of images. The different sets of data may consist of two or more digital images taken of a single scene from different sensors at different time or from different viewpoints. In image registration the geometric correspondence between the images is established so that they may be transformed, compared and analyzed in a common reference frame. This is of practical importance in many fields, including remote sensing, computer vision, medical imaging. Registration methods can be loosely divided into the following classes: algorithms that use image pixel values directly, e.g., correlation methods [2];algorithms that use the frequency domain, e.g., Fast Fourier transform based (FFT-based) methods [3];algorithms that use low level features such as edges and corners, e.g., Feature based methods [4];and algorithms that use high-level features such as identified parts of image objects, relations between image features, for e.g., Graph-theoretic methods[4].The next step, following registration, is image warping which includes correcting distorted images and it can also be used for creative purposes. The images are placed appropriately on the bigger canvas using registration transformations to get the output mosaiced image. The quality of the mosaiced image and the time efficiency of the algorithm used are given most importance in image mosaicing.

Image Blending is the technique which modifies the image gray levels in the vicinity of a boundary to obtain a smooth transition between images by removing these seams and creating a blended image. Blend modes are used to blend two layers into each other.

*A. Image Mosaicing Model*

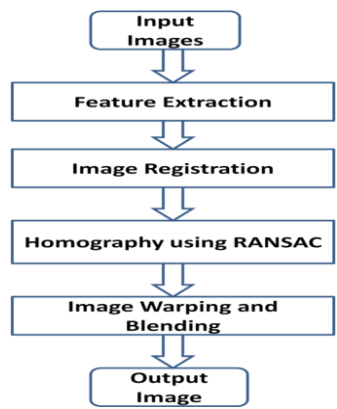

## II. FEATURE DETECTION

Feature detection is take place as a very first step in the process of Image Mosaicing. Features are the elements in the two input images to be matched. For images to be matched they are taken inside an image patches. These image patches are groups of pixel in images. Patch matching is done for the input images. It is clearly explained below as follows:-

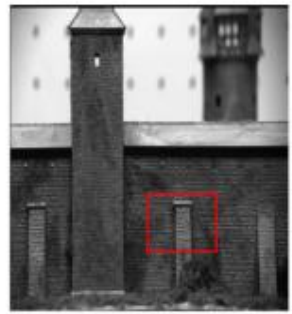
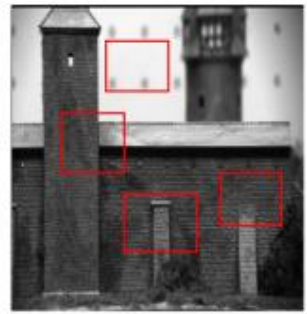

Patch-Matching.Fig-1: Input Image-1        Fig-2: Input Image-2

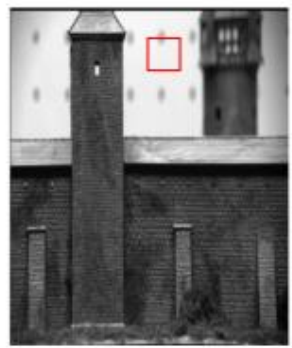
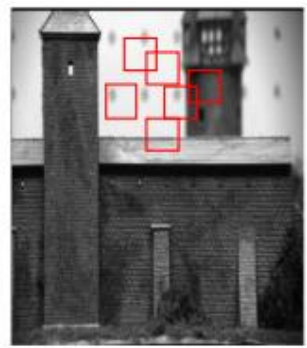

Patch-Matching.Fig-3: Input Image-1        Fig-4: Input Image-2

In the given example, if we notice then fig 1 and fig2 gives a best patch match as there is one patch in fig2 which is looking exactly similar to the patch which is given in fig1. When we consider fig3 and fig4, here it's a bad patch match as there are many similar patches in fig4 which are looking similar to the patch so given in fig3.So, exact feature matching cannot be done because intensities are slightly equal here.

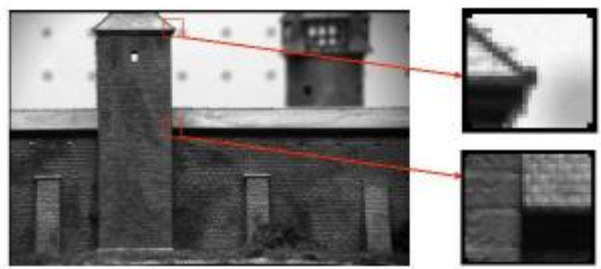

Fig-5: Corners (Junction of Contours.)

To provide a better feature matching for the pairs of images, corners are compared to give quantitative measurement. Corners are best features to match. The features of corners are that they are more stable features over changes of viewing angles. The other most important feature of corner is that if there is a corner in an image than its neighborhood will show an abrupt change in intensity. Corners can be detected in images using various corner detection algorithms. Some of the corner detection algorithms are Harris Corner detection Algorithm, Susan corner detection algorithm, SIFT corner detection algorithm (Scale Invariant Feature Transform), the machine learning based FAST algorithm, Speeded-up robust feature (SURF).

*A. Harris Corner Detection Algorithm*

This Algorithm was developed by Chris Harris and Mike Stephens in 1988 as a low level processing step to aid researchers trying to build interpretations of a robot's environment based on image sequences. Specifically, Harris and Stephens were interested in using motion analysis techniques to interpret the environment based on images from a single mobile camera. Like Moravec, they needed a method to match corresponding points in consecutive image frames, but were interested in tracking both corners and edges between frames. Harris and Stephens developed this combined corner and edge detector by addressing the limitations of the Moravec operator. The result was far more desirable detector in terms of detection and repeatability rate at the cost of requiring significantly more computation time. Despite the high computational demand, this algorithm was widely used in practice.

A local detecting window in image is designed. The average variation in intensity that results by shifting the window by a small amount in different direction is determined. At this point the center point of the window is extracted as corner point. We can easily get the point by looking at intensity values within a small window. Shifting the window in any direction gives a large change in appearance. Harris corner detector is used for corner detection. On shifting the window if it's a flat region than it will show no change of intensity in all direction. If an edge region is found than there will be no change of intensity along the edge direction. But if any corner is found than there will be a significant change of intensity in all direction. Harris corner detector gives a mathematical approach for determining whether the region is flat, or if there is an edge or corner. Harris corner technique is very much helpful in detecting more features and that technique is rotational invariant and scale variant. For the change of intensity for the shift [u, v].

$$E(u,v) = \sum_{x,y} w(x,y)[I(x+u, y+v) - I(x,y)]^2$$

Where w(x, y) is a window function, I(x + u, y + v) is the shifted intensity and I(x, y) is the intensity of the individual pixel. Harris corner algorithm is given below as:

1. For each pixel (x, y) in the image calculate the autocorrelation matrix M as;

$$M = \sum_{x,y} \begin{bmatrix} I_x^2 & I_x I_y \\ I_x I_y & I_y^2 \end{bmatrix}$$

2. For each pixel of image has Gaussian filtering, get new matrix M, and discrete two-dimensional zero-mean Gaussian function as:

$$\text{Gauss} = \exp(-u^2+v^2)/2\delta^2$$

3. Calculating the corners measure for each pixel (x, y,), we_get;

$$R = \{I_x^2 x I_y^2 - (I_x^2 I_y^2)\} - k\{I_x^2 + I_y^2\}^2$$

4. Choose the local maximum point. Harris method considers that the feature points are the pixel value which corresponding with the local maximum interest point.

5. Now Set the threshold T, and detect corner points.

*B. SIFT Algorithm*

SIFT Algorithm is Scale Invariant Feature Transform. SIFT is a corner detection algorithm which detects features in an image which can be used to identify similar objects in other images. SIFT produces key-point-descriptors which are the image features. When checking for an image match two set of key-point descriptors are given as an input to the Nearest Neighbor Search (NNS) and produces a closely matching key-point descriptors. SIFT has four computational phases which includes: Scale-space construction, Scale-space extreme detection, key-point localization, orientation assignment and defining key-point descriptors. The first phase identifies the potential interest points. It searches over all scales and image locations by using a difference-of Gaussian function. For all the interest points so found in phase one, location and scale is determined.

Key-points are selected based on their stability. A stable key-point should be resistant to image distortion. In Orientation assignment SIFT algorithm computes the direction of gradients around the stable key-points. One or more orientations are assigned to each key-point based on local image gradient directions. For a set of input frames SIFT extracts features. Image matching is done using Best Bin First (BBF) algorithm for estimating initial matching points between input frames. To remove the undesired corners which do not belong to the overlapped area, RANSAC algorithm is used. It removes the false matches in the image pairs. Reprojection of frames are done by defining its size, length, width. Stitching is done finally to obtain a final output mosaic image. In stitching, each pixel in every frame of the scene is checked whether it belongs to the warped second frame. If so, then that pixel is assigned the value of the corresponding pixel from the first frame. SIFT algorithm is both rotational invariant and scale invariant. SIFT is very suitable for object detection in images with high resolution.

It is also a robust algorithm for image comparison though it is comparatively slow. The running time of a SIFT algorithm is large as it takes more time to compare two images.

*C. FAST Algorithm*

FAST is a corner detector algorithm. Trajkovic and Hedley founded this algorithm in 1998.The detection of corner was prioritized over edges in FAST as corners were found to be the good features to be matched because it shows a two dimensional intensity change, and thus well distinguished from the neighboring points. According to Trajkovic and Hedley the corner detector should satisfy the following criteria:-

1. The detected positions should be consistent, insensitive to the variation of noise, and they should not move when multiple images are acquired of the same scene.

2. Accuracy; Corners should be detected as close as possible to the correct positions.

3. Speed; the corner detector should be fast enough. FAST incremented the computational speed required in the detection of corners. This corner detector uses a corner response function (CRF) that gives a numerical value for the corner strength based on the image intensities in the local neighborhood. CRF was computed over the image and corners which were treated as local maxima of the CRF. A multi-grid technique is used to improve the computational speed of the algorithm and also for the suppression of false corners being detected. FAST is an accurate and fast algorithm that yields good localization (positional accuracy) and high point reliability.

D. SURF Algorithm:

The Speed-up Robust Feature detector (SURF) uses three feature detection steps namely; Detection, Description and Matching. SURF speeded-up the SHIFT's detection process by keeping in view of the quality of the detected points. It gives more focus on speeding-up the matching step. The Hessian matrix is used along with descriptors low dimensionality to significantly increase the matching speed. SURF is widely used in the computer vision community. SURF has proven its efficiency and robustness in the invariant feature-localization.

### III. COMPUTING HOMOGRAPHY

A. RANSAC Algorithm:

Calculating Homography is the third step of Image mosaicing. In homography undesired corners which do not belong to the overlapping area are removed. RANSAC algorithm is used to perform homography. RANSAC is an abbreviation for "RANdom Sample Consensus." It is an iterative method to estimate parameters of a mathematical model from a set of observed data which contains outliers. It is a non-deterministic algorithm in the sense that it produces a reasonable result only with a certain probability, with this probability increasing as more iterations are allowed. The algorithm was first published by Fischler and Bolles. RANSAC algorithm is used for fitting of models in presence of many available data outliners in a robust manner. Given a fitting problem with parameters considering the following assumptions.

1. Parameters can be estimated from N data items.

2. Available data items are totally M.

3. The probability of a randomly selected data item being part of a good model is $P_g$.

4. The probability that the algorithm will exit without finding a good fit if one exists is $P_{fail}$.

Then, the algorithm:

1. Selects N data items at random.

2. Estimates parameter x.

3. Finds how many data items (of M) fit the model with parameter vector x within a user given tolerance. Call this K.

4. If K is big enough, accept fit and exit with success.

5. Repeat 1.4 L times.

6. *Fail if you get here.*

How big K has to be depends on what percentage of the data we think belongs to the structure being fit and how many structures we have in the image. If there are multiple structures than, after a successful fit, remove the fit data and redo RANSAC.

We can find L by the following formulae:
$P_{fail}$ = Probability of L consecutive failures.
$P_{fail}$ = (Probability that a given trial is a failure) L.
$P_{fail}$ = (1 - Probability that a given trial is a success)L.
$P_{fail}$ = (1-(Probability that a random data item fits the model)N)L.
$P_{fail} = (1-(P_g)^N)^L$
$L = \log(Pfail) / \log(1 - (Pg)N)$

*B. Homography:*

Homography is mapping between two spaces which often used to represent the correspondence between two images of the same scene. It's widely useful for images where multiple images are taken from a rotating camera having a fixed camera *Centre* ultimately warped together to produce a panoramic *view.*

## IV. IMAGE WARPING AND BLENDING

**Image warping** is the process of digitally processing a photo such that any shapes portrayed in the photo have been significantly distorted. Warping may be used for correcting photo distortion as well as for creative purposes.

The final step is to **blend** the pixels colors in the overlapped region to avoid the seams. Simplest available form is to use feathering, which uses weighted averaging color values to blend the overlapping pixels. We generally use alpha factor often called alpha channel having the value 1 at the center pixel and becomes 0 after decreasing linearly to the border pixels. Where at least two photos overlap occurs in an output mosaic we will use the alpha values as follows to compute the color at a pixel in there, suppose there are two photos, I1, I2, overlapping in the output photo; each pixel (x,y) in photo Ii is represented as Ii(x, y) = (αiR, αiG, αiB, αj) where (R,G,B) are the color values at the pixel. We will compute the pixel value of (x, y) in the stitched output photo as [(α1R, α1G, α1B, α1) + (α2R, α2G, α2B, α2)]/(α1+α2).

## V. IMAGE WARPING AND BLENDING

Image Mosaicing techniques are widely used in creating panoramic or stretching of images. In this paper, some of the popular algorithms have been explores. Harris corner detection method is robust as well as rotationally invariant. However, it is scale variant. The FAST algorithm is both rotation and scale invariant with optimized execution time. But, its performance is poor when noise is present. SIFT algorithm is rotation as well as scale invariant and more effective in presence of noise. It has highly distinctive features However, it suffers from illumination variation.

## VI. Conclusion

Image Mosaicing techniques are widely used in creating panoramic or streching of images. In this paper, some of the popular algorithms have been explores. Harris corner detection method is robust as well as rotationally invariant. However, it is scale variant. The FAST algorithm is both rotation and scale invariant with optimized execution time. But, its performance is poor when noise is present. SIFT algorithm is rotation as well as scale invariant and more effective in presence of noise. It has highly distinctive features However, it suffers from illumination variation.

**References**

1. Jain, D.K.; Saxena, G .;Singh, V.K., "Image mosaicing using corner technique" in International Conference on Communication System and Network Technologies, pp 79-84, 2012.
2. D. I. Barnea; and H. F. Silverman,"A class of algorithms for fast digital registration," IEEE Trans. Comput, vol.C-21, pp.179-186, 1972.
3. C. D. Kuglin and D .C. Hines," The phase correlation image alignment method",in Proc. IEEE Int. Conf. Cybernet. Society, New York, NY, pp 163-165, 1975.
4. Lisa G. Brown. A survey of image registration techniques. ACM Computing Surveys, 24(4 ); pp 325-376, December 1992.
5. Jinpeng. Wang.; Yewei Li, " Image mosaicing algorithm based on salient region and MVSC " in International Conference on Multimedia and Signal Processing, pp 207-211, 2011.
6. Ghosh,D.; Park,S.;Kaabouch,N.;Semke,W." Quantum evaluation of image mosaicing in multiple scene categories "IEEE Conference on Electro/Information Technology,pp 1-6,2012.
7. Azzari,P.; Stefano,L.D.; Stefano,M., Stefano,M.," An evaluation methodology for image mosaicing algorithms" Advanced Concept for Intelligent Vision Systems,Springer,vol.5259 ,pp 89-100,2008.
8. Brunet.Florent; &Gay-Bellile.Vincet;et al."Feature-Driven Direct Non-Rigid Image Registration." International Journal of Computer Vision, 2011, vol93, pp : 33-52.
9. Zhiqian.Y;&Hao.W;et al.,"An Image Mosaic Algorithm of Pathological section based on feature points."International Conference on Information Engineering and Computer Science, 2009, pp. 1-3.
10. Peleg,S.:Herman,J.,"Panoramic mosaics by manifold projection" IEEE Computer society conference on Computer Vision and Pattern Recognition,pp 338-343,1997.
11. Rousso,B.;Peleg,S.;Finci,I.;Rav-Acha,A.," Universal mosaicing using pipe projection"Sixth International Conference on Computer Vision, pp 945-950,1998.
12. Ward, Greg. "Hiding seams in high dynamic range panoramas". Proceedings of the 3rd symposium on Applied perception in graphics and visualization.. ACM International Conference Proceeding Series,pp 153-155,2006.
13. Wolberg, G. Digital Image Warping, IEEE Computer Society Press,pp 169-172 ,1990.
14. Inampudi,R.B.," Image mosaicing" in International Conference on Geoscience and Remote Sensing Symposium Proceedings,vol.5,pp 2363-2365,1998

# Author Profile

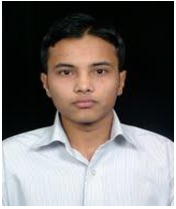
**Dushyant Vaghela,** Received the BE degree in Information Technology from Saurashtra Univesity and Pursing M.Tech degree in Computer Engineering from RK University, He is the webmaster at explorequotes web, expert in SEO, SMO and Publisher at Google Adsense.

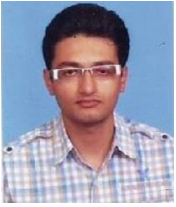
**Kapildec Naina,** Assistant professor at RK University in Computer Engineering Department, Specialist in Digital Image Processing, and computer graphics.